\documentclass{article}

    \PassOptionsToPackage{numbers, compress, square}{natbib}

\usepackage[preprint]{neurips_2022}

\usepackage[utf8]{inputenc} 
\usepackage[T1]{fontenc}    
\usepackage{hyperref}       
\usepackage{url}            
\usepackage{booktabs}       
\usepackage{amsfonts}       
\usepackage{nicefrac}       
\usepackage{microtype}      
\usepackage{xcolor}         
\usepackage{todonotes}      
\usepackage{graphicx}       
\usepackage{hyperref}
\usepackage[toc,page]{appendix}
\usepackage{subcaption}
\usepackage{setspace}

\usepackage{comment}

\title{On The Fragility of Learned Reward Functions}

\author{
  Lev McKinney\footnotemark[1]\\
  University of Toronto\\
  \texttt{lev.mckinney@mail.utoronto.ca}\\
  \And
  Yawen Duan\footnotemark[1]\\
  University of Cambridge\\
  \texttt{yd338@cam.ac.uk}
  \And
  David Krueger\\
  University of Cambridge\\
  \texttt{david.scott.krueger@gmail.com}\\
  \And
  Adam Gleave\\
  University of California, Berkeley\\
  \texttt{gleave@berkeley.edu}\\
}

\newif\iflongversion
\longversionfalse 
\begin{document}
\maketitle

\footnotetext[1]{Equal contribution. Work done during internship at Center for Human-Compatible AI, UC Berkeley.}

\begin{abstract}
Reward functions are notoriously difficult to specify, especially for tasks with complex goals. Reward learning approaches attempt to infer reward functions from human feedback and preferences. Prior works on reward learning have mainly focused on the performance of policies trained alongside the reward function. This practice, however, may fail to detect learned rewards that are not capable of training new policies from scratch and thus do not capture the intended behavior. Our work focuses on demonstrating and studying the causes of these \textit{relearning} failures in the domain of preference-based reward learning. We demonstrate with experiments in tabular and continuous control environments that the severity of relearning failures can be sensitive to changes in reward model design and the trajectory dataset composition. Based on our findings, we emphasize the need for more retraining-based evaluations in the literature.
\end{abstract}
\section{Introduction}

Reward functions for most real-world tasks are difficult or impossible to specify procedurally.
Specifically, hand-designed reward functions frequently misspecify the task~\citep{krakovnaSpecificationGaming2020}.
The field of reward learning attempts to overcome this challenge by designing algorithms to infer reward functions from data. These learned reward functions aim to succinctly represent the desired behaviors \citep{ng2000algorithms}, drastically reduce the amount of human feedback required to learn a task \citep{christianoDeepReinforcementLearning2017} and allow practitioners to generalize these behaviors to new environments \citep{fuLearningRobustRewards2018}.

\iflongversion
\else {
One of the most promising approaches is to learn reward functions from binary human preferences over trajectory segments where these segments are collected online using a \emph{sampler agent} trained to optimize the learned reward \cite{christianoDeepReinforcementLearning2017}. This form of preference-based reward learning is already being used to train 
large language models to summarize \citep{ZieglerFineTuning2019} and become more helpful and harmless \citep{baiTrainingHelpfulHarmless2022}.
}\fi

Prior work has typically focused on the performance of the sampler agent \cite{leePEBBLEFeedbackEfficientInteractive2021, christianoDeepReinforcementLearning2017}. Unfortunately, the sampler agent performing the correct behavior does not guarantee that a robust reward function has been uncovered. In particular, when using reinforcement learning to train a randomly initialized \emph{relearner agent} on the learned reward, the reward may fail to induce the correct behavior despite the sampler agent behaving well \cite{ibarzRewardLearningHuman2018}. If we only require a policy that works passably well in the exact training environment, this may not be an issue because we can use the sampler agent and throw away the learned reward. We argue, however, that such a method cannot be accurately described as learning a \emph{reward} function. At most, it is a preference-based policy learning technique, using reward functions to give a helpful inductive bias during training. Moreover, it is desirable for many applications to truly uncover a reward function. For example, we might wish to train a new policy using a learned reward function with a more powerful R.L. algorithm or different agent architecture than was used during the initial reward learning process.

Past work has preformed preliminary investigations into the robustness of learned rewards in toy environments \citep{reddyLearningHumanObjectives2021}, in Atari \citep{ibarzRewardLearningHuman2018} and for fine-tuning language models \citep{baiTrainingHelpfulHarmless2022, stiennonLearningSummarizeHuman2022}. However, these investigations are typically only reported short sections of their respective papers.

Inspired by this work, our paper empirically examines the relearner performance for learned reward functions. Since we have access to a ground truth reward in our synthetic experiments, we define poor relearner performance as achieving relatively low ground truth returns. Our results show that relearing can produce very different policies than the sampler, frequently achieving low ground-truth returns. Thus, we argue that current preference-based reward learning methods may produce reward functions that are not reliable as signals for policy relearning.

Our paper makes three key contributions:
\begin{itemize}
    \item We demonstrate that state-of-the-art reward learning algorithms can produce reward models that fail to train new agents from scratch in tabular and continuous control settings;
    \item We show that the severity can increase as the trajectory dataset concentrates on high reward regions;
    \item Finally, as an example of how these relearning failures can be sensitive to changes in reward model design, we demonstrate that reward ensembles can effect relearning failures.
\end{itemize}
 
\section{Related Work}
\label{relatedwork}
\paragraph{Preference-based reward learning} 

Our primary focus is on methods that learn from \textit{preference comparisons} between two trajectories~\citep{akrourPreference2011,wilsonPreference2012,sadighActive2017,christianoDeepReinforcementLearning2017}.
Preference comparison is one of the most scalable reward learning methods, successfully applied to fine-tune large transformer language models~\cite{ouyangTrainingLanguageModels2022, stiennonLearningSummarizeHuman2022, nakanoWebGPTBrowserassistedQuestionanswering2022,baiTrainingHelpfulHarmless2022} to enhance their performance at certain tasks. Note that trajectory comparison methods contain more information about the reward than demonstrations, so they tend to produce better results when available~\citep{skalseInvariance2022}. However, note that these methods may still fare poorly when the human preference feedback does not match their model of human rationality \cite{leeBPrefBenchmarkingPreferenceBased2021}.

\paragraph{Other reward learning approaches}
Many other methods have been developed to learn reward functions from human data~\citep{jeonRewardRational2020}.
One of the most popular is \emph{Inverse reinforcement learning} (IRL)~\citep{ng2000algorithms} methods that infer a reward function from demonstrations~\citep{abbeelApprenticeship2004,ramachandranBIRL2007,ziebartMaximum2008,ziebartModeling2010,ziebartPaper2010,finnGuidedCostLearning2016,fuLearningRobustRewards2018}.
\mbox{T-REX}~\citep{brownExtrapolating2019} is a hybrid approach, learning from a \textit{ranked} set of demonstrations.
An alternative approach learns from  ``sketches'' of cumulative reward over an episode\citet{cabiScalingDatadrivenRobotics2020}.

\paragraph{Reward hacking} 

\citeauthor{panEffectsRewardMisspecification2022} provides the first systematic empirical study of \emph{reward hacking}: RL agents exploiting misspecified reward functions~\citep{panEffectsRewardMisspecification2022}.
Notably, they find that increasing agent capabilities, such as by increasing the RL policy's model size, can sometimes lead to \emph{worse} performance on the ground truth reward, while performance on the misspecified \emph{proxy} reward increases. In contrast to our work, \citeauthor{panEffectsRewardMisspecification2022} only study reward hacking in hand-designed rewards designed to illustrate the phenomenon, whereas we investigate this phenomenon in learned rewards.

Reward hacking has also been studied from a theoretical perspective. Under the framework of general principle agent problems \citeauthor{zhuangConsequencesMisalignedAI2021} examines the case where the agent's utility function can only account for a limited subspace of the set of attributes that make the true utility~\cite{zhuangConsequencesMisalignedAI2021}. The authors proceed to show that, within their model, an optimal state under this proxy utility can have arbitrarily low ground truth utility, assuming the attributes that make up the reward exhibit a condition analogous to decreasing marginal utility and increasing opportunity cost. \citeauthor{skalseHacking2022} instead propose a formal definition of reward hacking~\citep{skalseHacking2022}.
In our paper, however, we focus on more concrete cases of relearning failures and practically attainable measures, such as relearner performance being close to sampler performance.

The most closely related work is by \citeauthor{ibarzRewardLearningHuman2018}, which evaluates their learned reward functions by freezing them and training a new policy, analogous to our \emph{relearner} evaluations~\citep[Section~3.2]{ibarzRewardLearningHuman2018}. However, this study was only a small, half-page section of their paper, and they did not examine factors that may increase or decrease the chances/severity of relearning failures. 

Another important related work is by \citeauthor{reddyLearningHumanObjectives2021}, who observes that rewards can fail to generalize due to a lack of informative trajectories in their training data ~ \citep{reddyLearningHumanObjectives2021}. They attempt to ameliorate this by querying humans on diverse hypothetical trajectories generated from a model. However, their method requires a world model and primarily focuses on taking advantage of this model to improve reward quality. In addition, they assume the user provides feedback through quantitative reward labels, whereas we focus on the more realistic and widely used preference comparison setting.

Finally, past work has found that language models trained on a preference-based reward model can learn to exploit their reward model~\citep[Section~4.3]{stiennonLearningSummarizeHuman2022}. In a similar vein, \citeauthor{baiTrainingHelpfulHarmless2022} found that the ability of their reward model to correctly predict human preferences over a pair of inputs degraded as those inputs where perceived as more rewarding by the model ~ \citep[Section~4.2]{baiTrainingHelpfulHarmless2022}. However, none of these works have offered much analysis of what leads to reward hacking or in general relearning failures, beyond training against the learned reward for to long \citep[Section~4.3]{stiennonLearningSummarizeHuman2022} or a lack of data from off distribution \cite{reddyLearningHumanObjectives2021}.

\paragraph*{Retraining and transfer in IRL domain} There has also been multiple works exploring relearning and transfer when learning rewards learned from expert demonstrations i.e. inverse reinforcement learning (IRL).
\citeauthor{fuLearningRobustRewards2018}~\citep{fuLearningRobustRewards2018} propose an IRL method to learn state-only reward functions disentangled from transition dynamics and preform experiments on transferring their learned rewards to new agents and environments. \citeauthor{niFIRLInverseReinforcement2020}~\citep{niFIRLInverseReinforcement2020} derive an analytic gradient estimator for an arbitrary f-divergence between expert and on policy distributions with respect to the reward functions parameters. In their relearning evaluations, they find that there method produces relearners that match expert performance.
Finally, \citeauthor{wangRandomExpertDistillation2019}~\citep{wangRandomExpertDistillation2019} borrows methods from random network distillation to directly estimate the expert distribution with only expert data. This process, removes the need for a sampler, obviating the issue of relearning failures. In contrast to these IRL methods, our work focuses on the more scalable preference-based reward learning setting.

\section{Background}
\label{sec:background}

\paragraph{Deep RL from Human Preferences}


We follow the framework of learning a preference model $\hat r_\phi$ from trajectory segment comparisons. Our method is the closest to deep reinforcement learning from human preferences \cite{christianoDeepReinforcementLearning2017}. It consists of four phases iterated: trajectory collection, preference elicitation, reward inference and policy optimization. During \textbf{trajectory collection}, the current policy, initially a random policy, samples rollouts from the environment collecting trajectory segments $\sigma_i = (s_0,a_0,s_1,a_1,\cdots,s_n)$ without reward labels and stores them in $\mathcal{B}$. In phase two, the algorithm, \textbf{elicits preferences} $y \in \{\succ, \prec, \equiv\}$ for randomly selected pairs of segments $(\sigma_1, \sigma_2) \in \mathcal{B}$ from a labeler --- human or synthetic \footnote{We follow \citeauthor{ibarzRewardLearningHuman2018} in selecting preference pairs to query uniformly at random \citep{ibarzRewardLearningHuman2018}}. The preferences are then stored in a preference dataset $\mathcal{D}$. The algorithm assumes these preferences have been sampled from the Bradly-Terry model \cite{bradleyRankAnalysisIncomplete1952},

\begin{equation}
\label{eq:bradley-terry}
P(\sigma_1 \succ \sigma_2) = \frac{\exp\left(\sum_{s, a,s'\in \sigma_1} r(s, a, s')\right)}{\exp\left(\sum_{s, a,s'\in \sigma_1} r(s, a, s')\right) + \exp\left(\sum_{s, a, s' \in \sigma_2} r(s, a, s')\right)}.
\end{equation}
a widely used approximate model for human data in the preference based reward learning literature \citep{christianoDeepReinforcementLearning2017, ibarzRewardLearningHuman2018, leePEBBLEFeedbackEfficientInteractive2021}. In the third phase \textbf{reward inference}, the reward $\hat r_\phi$ is fit by using Adam \citep{kingma2017adam} to minimize the negative log likelihood of $\hat r_\phi$ under $\mathcal{D}$. The fourth and final phase of each iteration consists of \textbf{policy optimization}. In this stage, we can apply existing deep reinforcement learning algorithms to improve our policies expected return under the learned reward, and the process repeats.

%


\section{Training and Evaluation Procedure}
\paragraph{Reward Learning}
We train reward models with \textbf{synthetic data} that is sampled from the Bradley-Terry model of Eq.~\ref{eq:bradley-terry} with $r$ set to the ground truth reward. In the tabular setting, we train the sampler policy using soft Q-Learning \cite{haarnojaReinforcementLearningDeep2017} and the learned reward networks simply take the current state as a one-hot vector for input. In the continuous control setting, we use soft actor-critic (SAC) \citep{haarnoja2018sacapps} from Stable-Baselines3 ~\citep{stable-baselines3} and the learned reward networks receive the observation, action and next observation as input. \iflongversion
\else {See Appendix \ref{appendix:hyperparameters} for further details.} \fi




\paragraph{Reward Ensembles} \citeauthor{christianoDeepReinforcementLearning2017} and \citeauthor{leePEBBLEFeedbackEfficientInteractive2021} use reward ensembles to estimate the uncertainty of the learned reward \citep{christianoDeepReinforcementLearning2017, leePEBBLEFeedbackEfficientInteractive2021}. We explore how these ensembles may have another benefit, reducing the variance of off-distribution transitions. As in prior works, we train each ensemble member on bootstrapped datasets, normalize their outputs separately and use their mean as the reward.

\paragraph{Relearning} We freeze the learned reward and train a new, randomly initialized,  \emph{relearner} policy to evaluate our reward functions. We evaluate this policy under the ground truth reward. This is similar to the method employed by \citeauthor{ibarzRewardLearningHuman2018} \cite[Section~3.2]{ibarzRewardLearningHuman2018} to study reward hacking. In the continuous control settings this consist of training a new agent from the learned reward function using the same R.L. algorithm as the sampler, then evaluating it under the ground truth reward. In the tabular setting, we simply solve for the soft-optimal policy ($\alpha=0.1$) \cite{haarnojaReinforcementLearningDeep2017} under the learned reward function. 
\section{Experiments}
First, we investigate the occurrence of relearning failures in the continuous control domain. We use \texttt{HalfCheetah} environment as our test bed since it has been used in past works on preference-based reward learning \citep{christianoDeepReinforcementLearning2017}. Here we find that increasing the number of training timesteps the sampler takes between sampling trajectories for labeling increases the severity of relearning failures. Next, we focus on the effects of reward model design and observe that reward ensembles may reduce reward hacking in tabular environments by reducing the variance of off-distribution transitions. To demonstrate this failure mode we use the stay inside environment which consists of two rooms separated by a wall with a small doorway. The agent receives reward for staying in the inside room see \autoref{fig:stay-in-gt}.

\subsection{Preference Trajectory Dataset Imbalance and Relearning Failure}
\iflongversion
\else {
\begin{figure}
    \vspace{-0.5cm}
    \begin{centering}
    \begin{tabular}{lll}
         \includegraphics[trim=0 0 0 0, clip, width=0.32\textwidth]{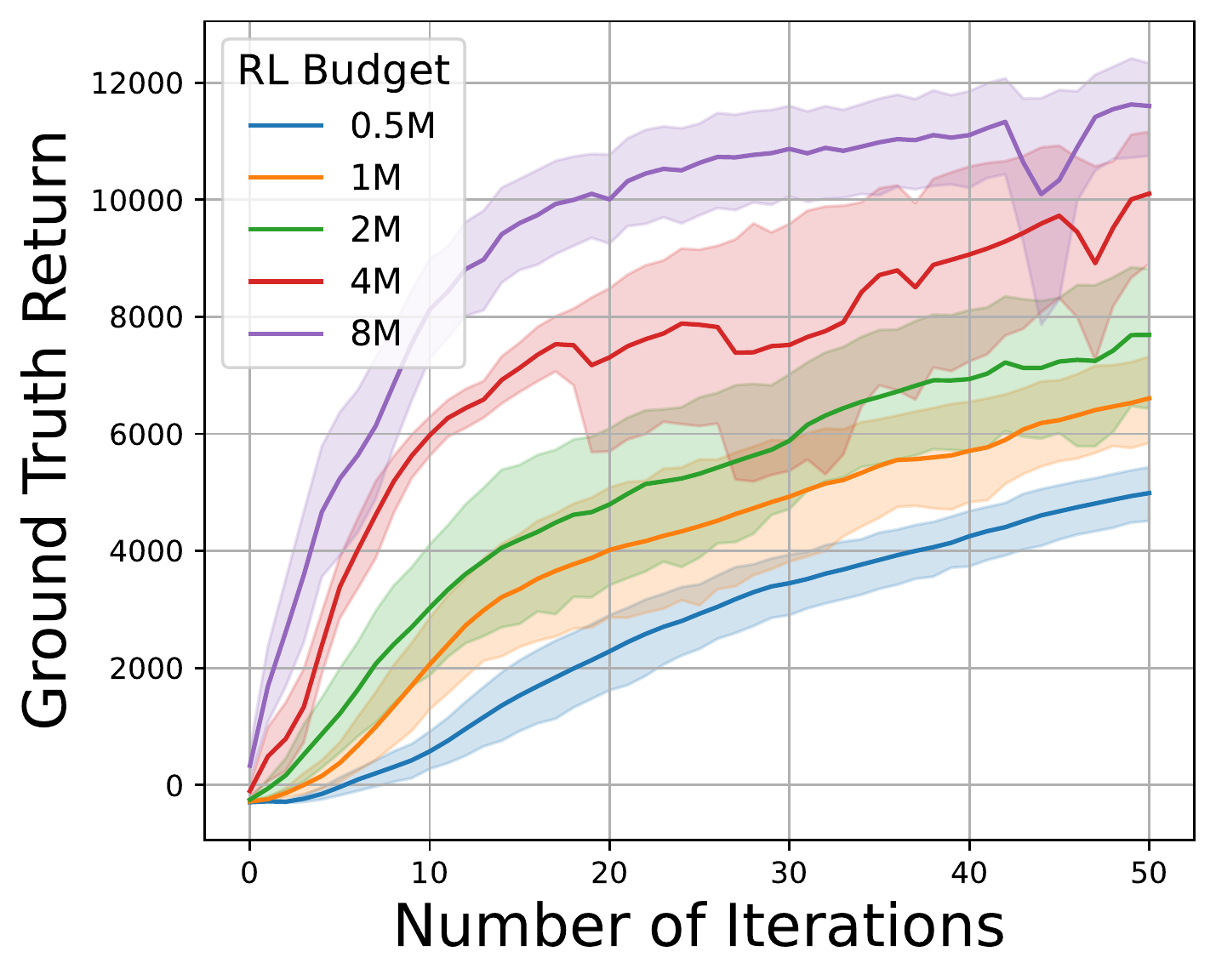}&
         \includegraphics[trim=0 0 0 0, clip, width=0.32\textwidth]{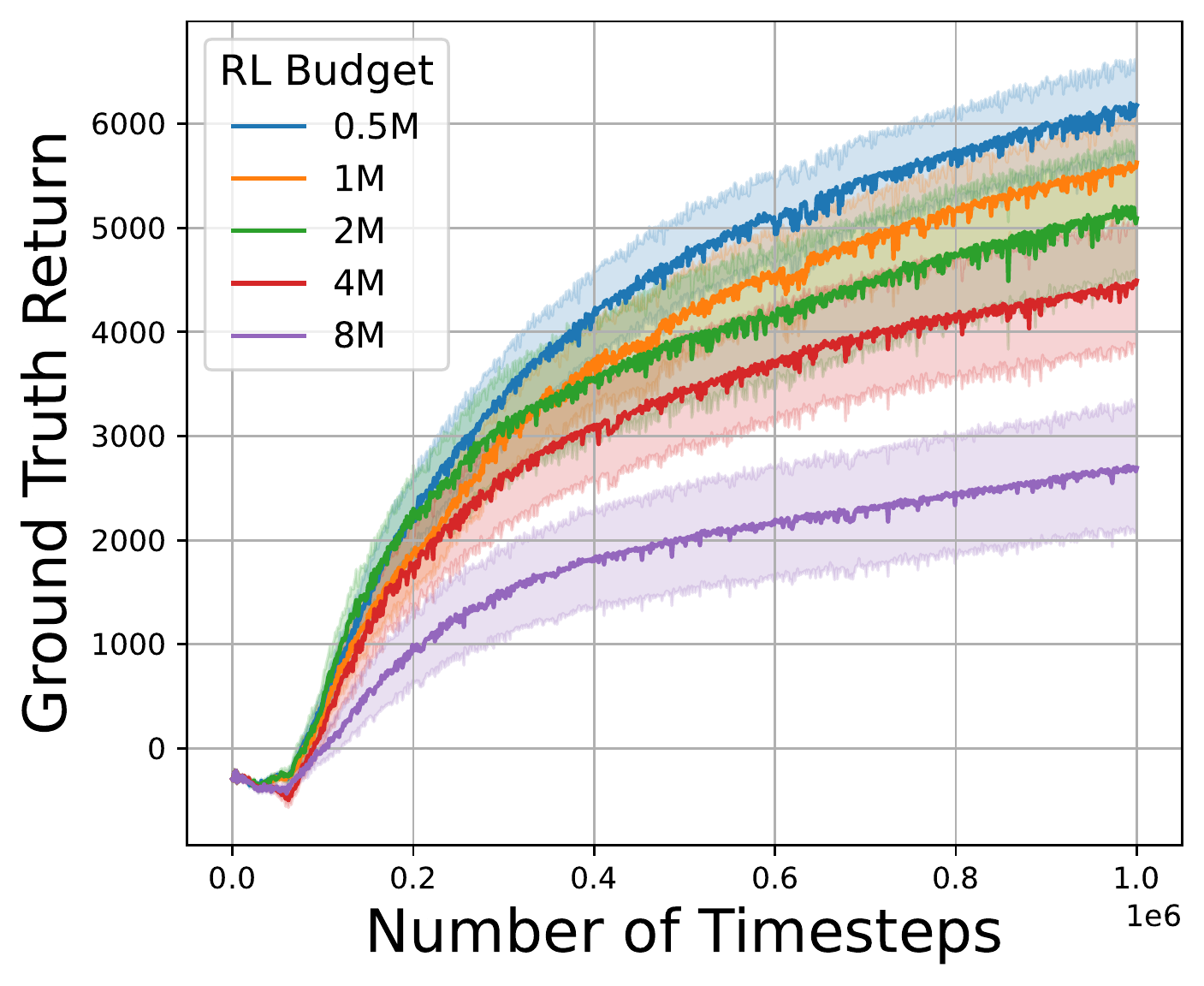}&
         \includegraphics[trim=0 0 0 0, clip, width=0.31\textwidth]{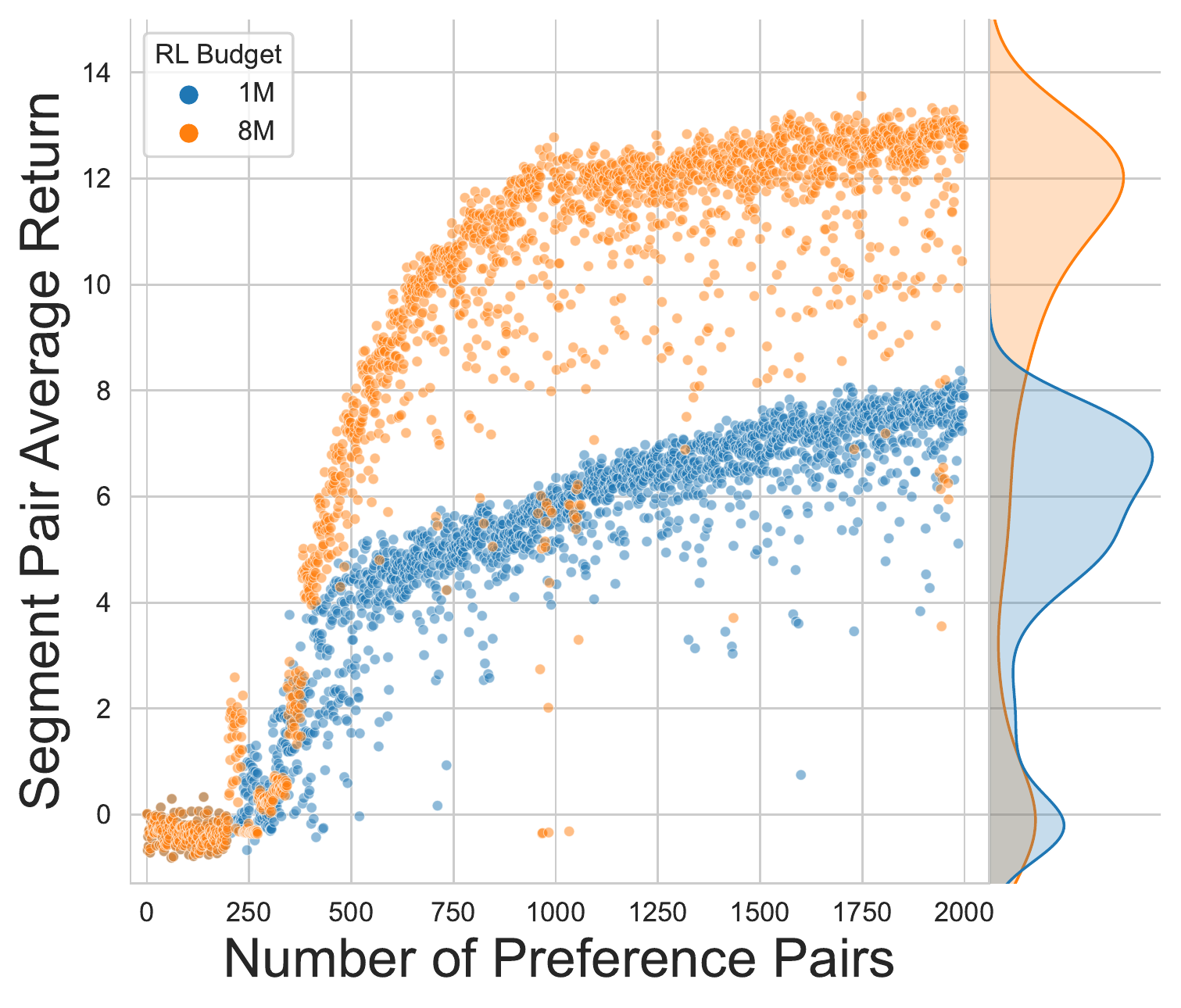}\\
         (a) \textit{Reward learning} curves &
         (b) \textit{Relearning} curves &
         (c) Preference datasets
    \end{tabular}
    \par\end{centering}
    \caption{Anti-correlated sampeler and relearner ground truth returns in \texttt{HalfCheetah}. (a) x-axis represents the number of iterations of each run. See section \ref{sec:background} RL budget is the total number of RL timesteps available to the sampler. (b) x-axis represents the number of timesteps during \textit{relearning}. In plots (a-b), for each RL budget setting, we performed ten runs of reward learning, and for each of these, we ran five relearning evaluations for a total of 50 relearning runs. Solid lines and shaded lines represent the mean and 90\% confidence respectively. (c) Scatterplot of average ground truth reward of each segment pair in the example preference datasets with 1M and 8M RL budgets.}
    \label{fig:longer}
    \vspace{-0.5cm}
\end{figure}



} \fi

The reward model is a function of the dataset $\mathcal{D}$ used to train it. One of the simplest ways to change the preference dataset is to vary the number of timesteps $T$ spent training the sampler between collecting trajectory fragments. We call the total number of interactions the sampler has with the environment during reward learning the \emph{RL budget}. Note the RL budget does \textbf{not} affect the number of comparisons collected.

Figure \ref{fig:longer} shows the learning curves of the sampler and relearner experiments in \texttt{HalfCheetah}. We find that despite higher RL budget leading to higher sampler returns during reward learning, the relearners' performance has the reverse trend; increasing sampler RL budget actually decreases relearner ground truth return.

We can gain some insight into why this is happening by exploring preference datasets shown in \iflongversion\else{\autoref{fig:reward_structure}c}\fi. First let's consider the preference dateset produced by one of the runs with the highest-budget (8M timesteps). We find that the trajectory segments contained in this datset are concentrated in high ground truth reward regions. On the other hand, when we consider the low-budget dataset (1M timesteps), the distribution of trajectory segments provides a better coverage across all the ground truth reward scales within the support.
\iflongversion
\else {
}\fi
We hypothesize that having an overwhelming proportion of high-reward trajectory segments in the preference dataset --- and little preference data on trajectories in the transition from high to low reward --- may cause the reward model to effectively over-fit to the high-reward region. This overfitting leads to poor supervision over randomly initialized policies. Overall, we believe this could explain the observed relearning failures.

It's important to note that we did not see a significant increase in relearning failure when increasing the RL budget in the tabular setting. See Appendix \ref{appendix:addition_tabular}.

\subsection{Reward Ensembles}
\iflongversion
\else {
\begin{figure}[!h]
    \vspace{-0.5cm}
    \begin{centering}
    \begin{minipage}{0.45\textwidth}%
    \begin{subfigure}{\linewidth}%
        \includegraphics[trim=0 5 15 40, clip, width=\textwidth]{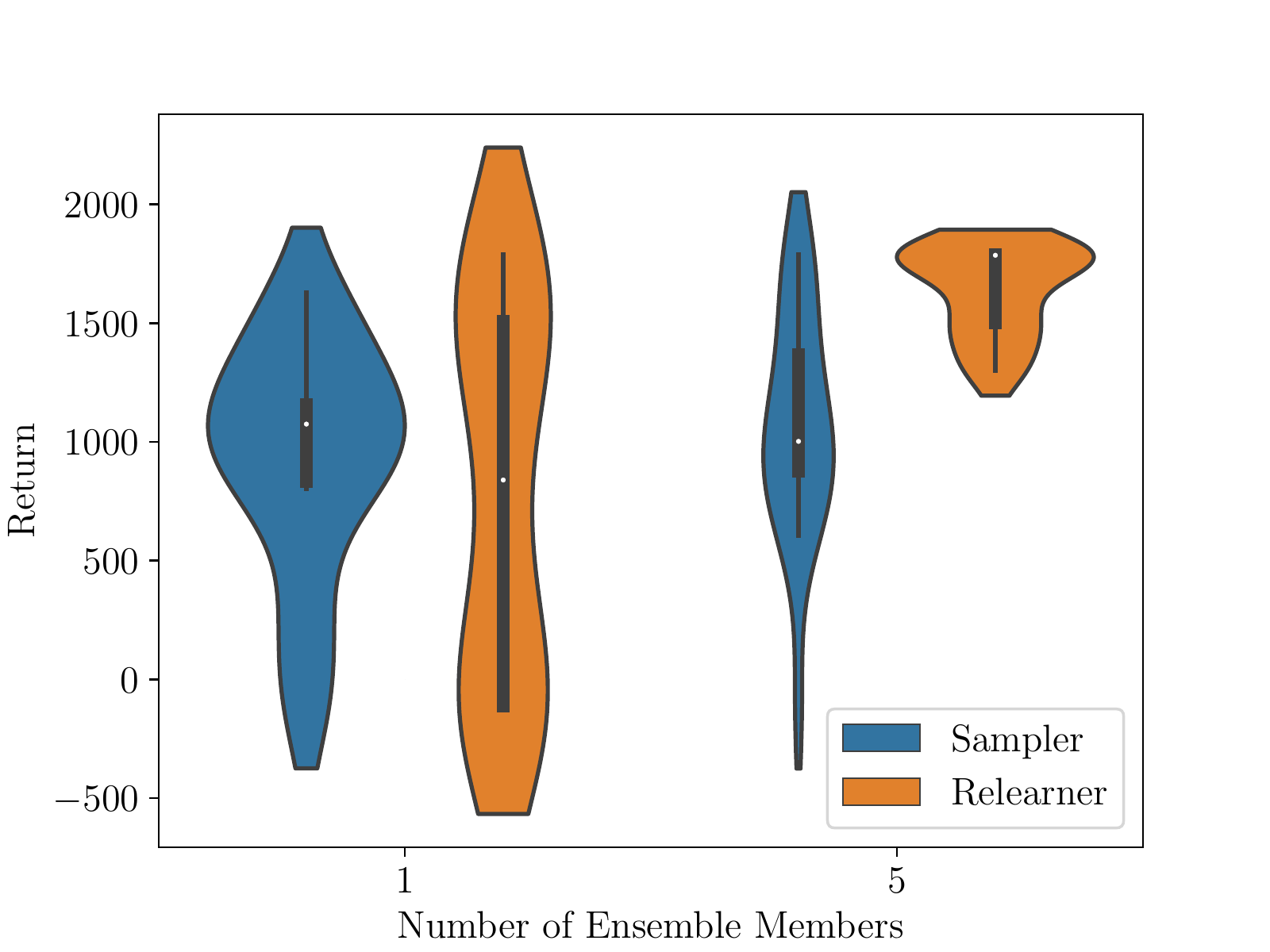}
        \caption{Effect of reward ensembles on sampler and relearner returns}
        \label{fig:relearner-and-sampler}
    \end{subfigure}
    \end{minipage}
    \begin{minipage}{0.25\textwidth}%
        \begin{minipage}{\textwidth}%
            \begin{subfigure}{\linewidth}%
                \centering
                \includegraphics[trim=10 0 10 0, clip,width=0.7\linewidth]{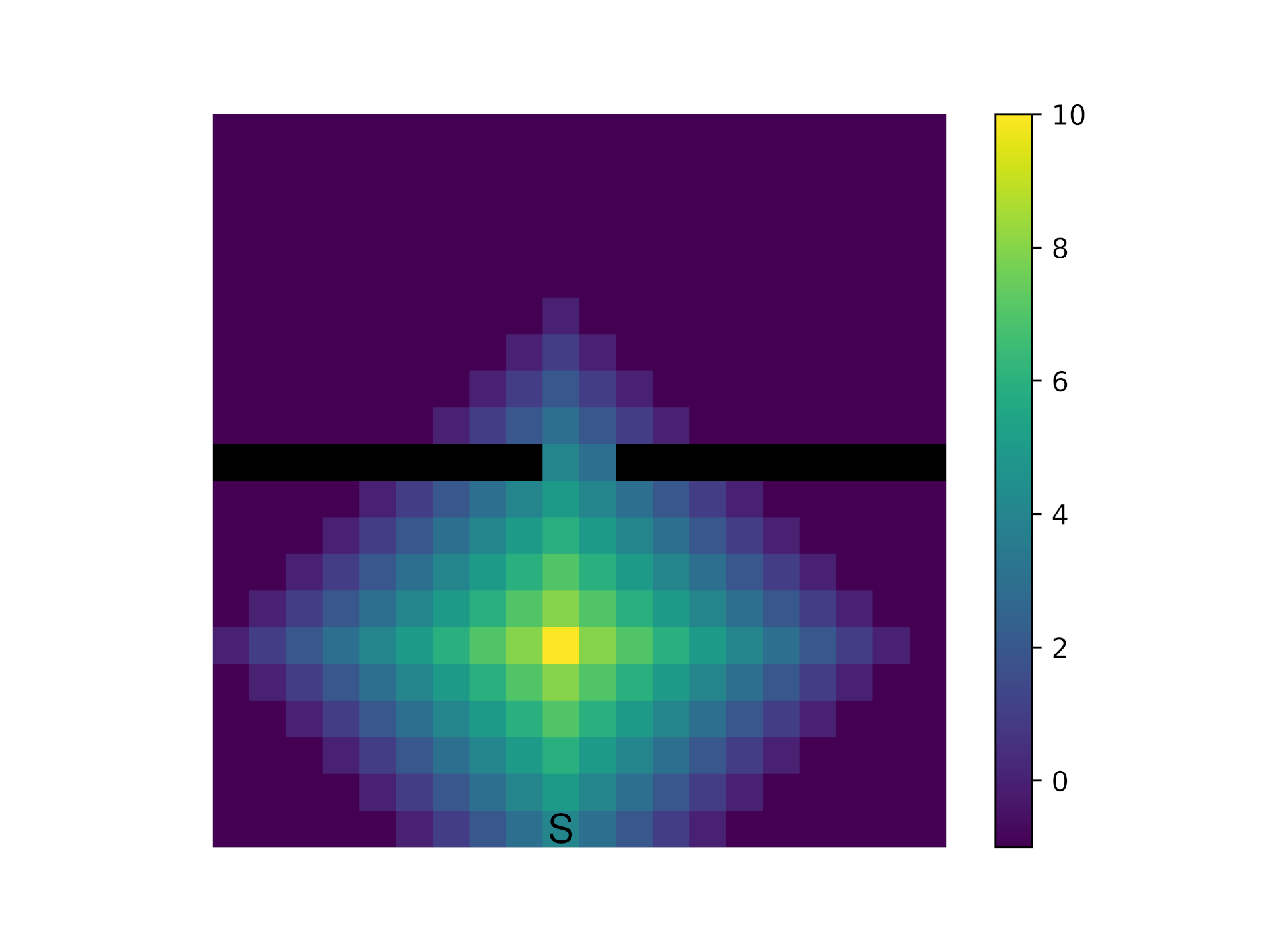}
                \caption{Ground truth reward} 
                \label{fig:stay-in-gt}
            \end{subfigure}
        \end{minipage}\\
        \begin{minipage}{\textwidth}%
            \begin{subfigure}{\linewidth}%
                \centering
                \includegraphics[trim=10 0 10 0, clip, width=0.7\linewidth]{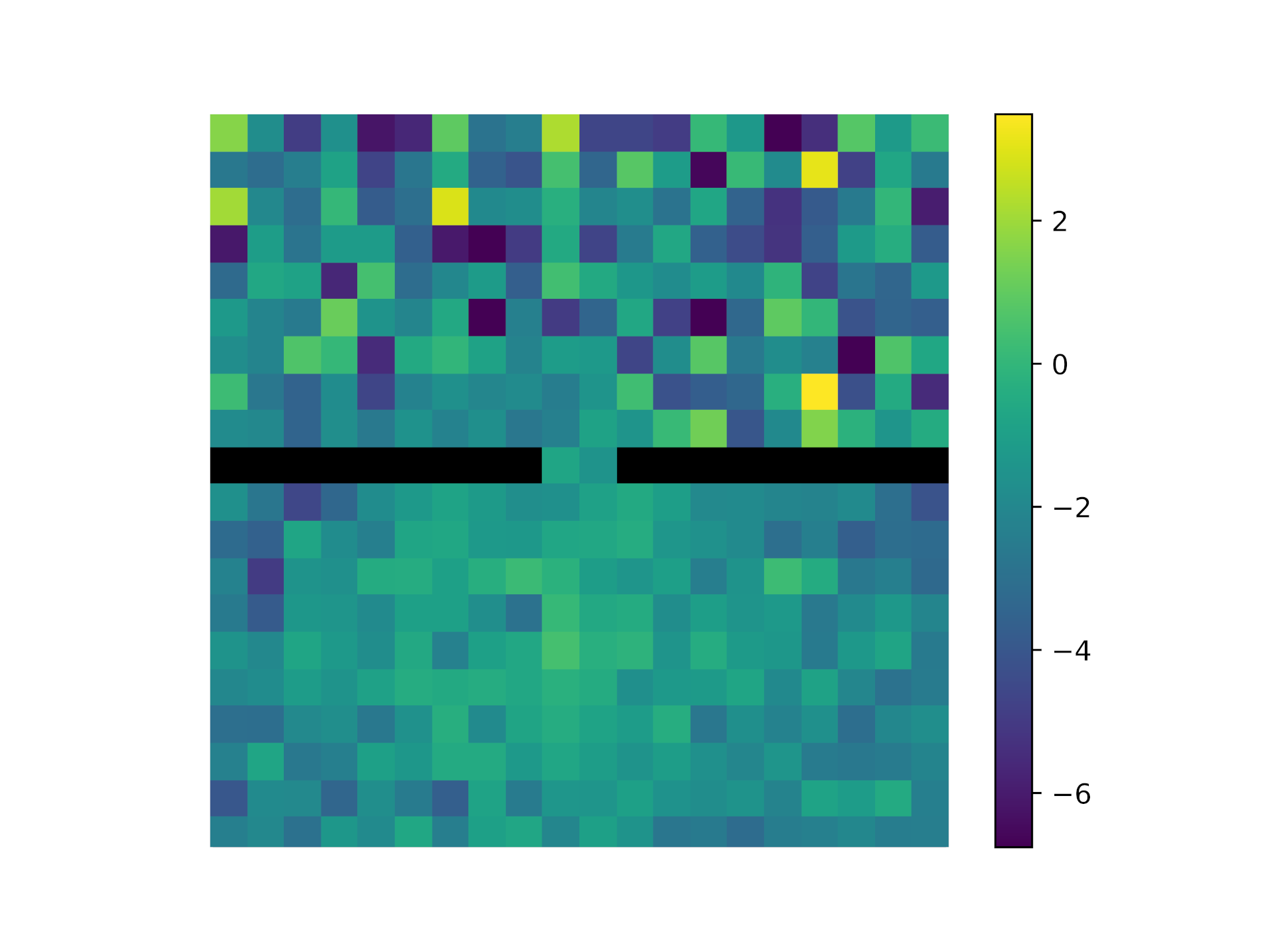}
                \caption{Example no ensemble} 
                \label{fig:stay-in-no-ensemlbe}
            \end{subfigure}
        \end{minipage}
    \end{minipage}
    \begin{minipage}{0.25\textwidth}%
        \begin{minipage}{\textwidth}%
            \begin{subfigure}{\linewidth}%
                \centering
                \includegraphics[trim=10 0 10 0, clip, width=.7\linewidth]{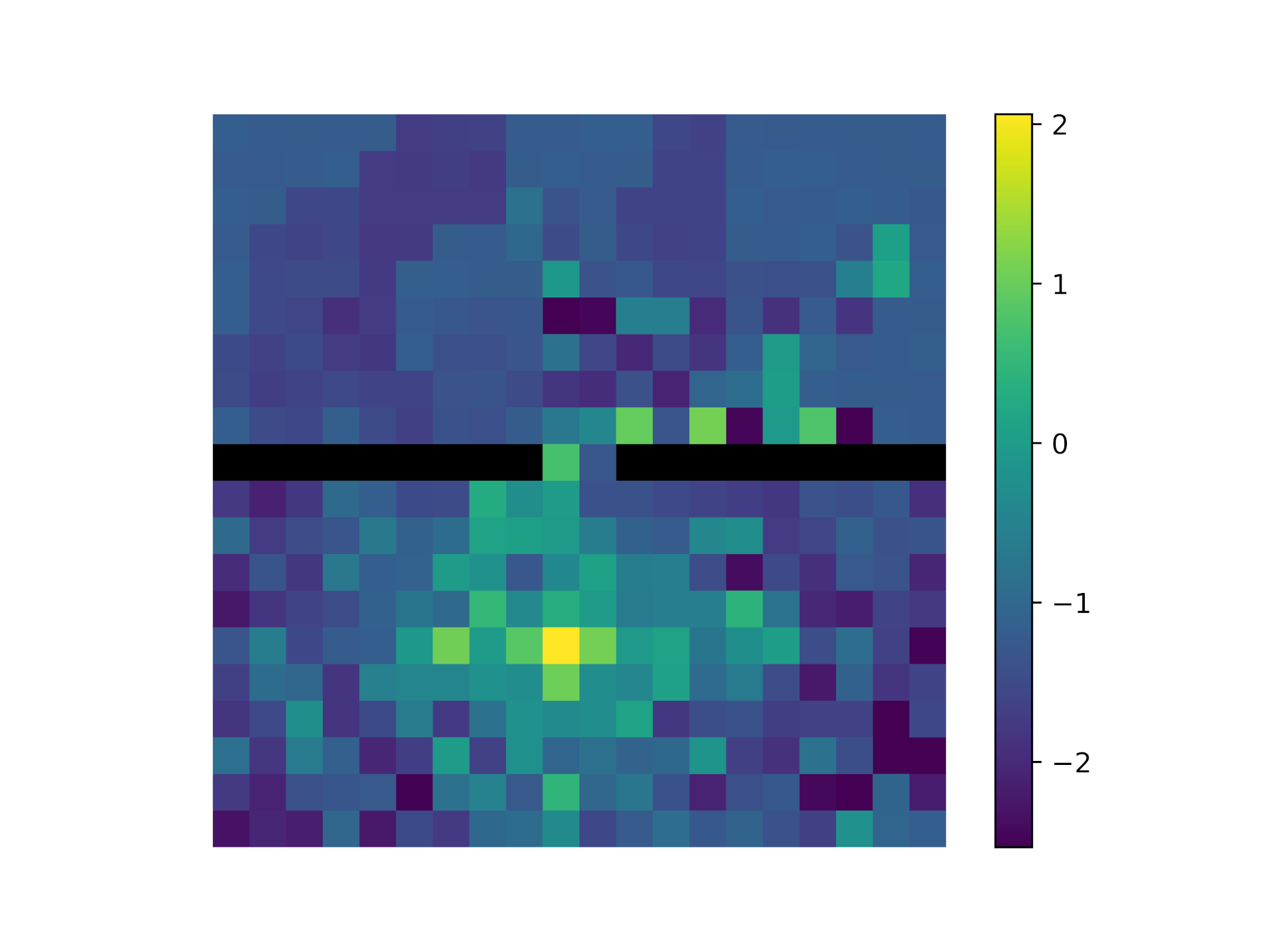}
                \caption{Example with ensemble} 
                \label{fig:stay-in-with-ensemble}
            \end{subfigure}
        \end{minipage}
        \begin{minipage}{\textwidth}%
            \begin{subfigure}{\linewidth}%
                \centering
                \includegraphics[trim=25 20 25 0, clip, width=1\linewidth]{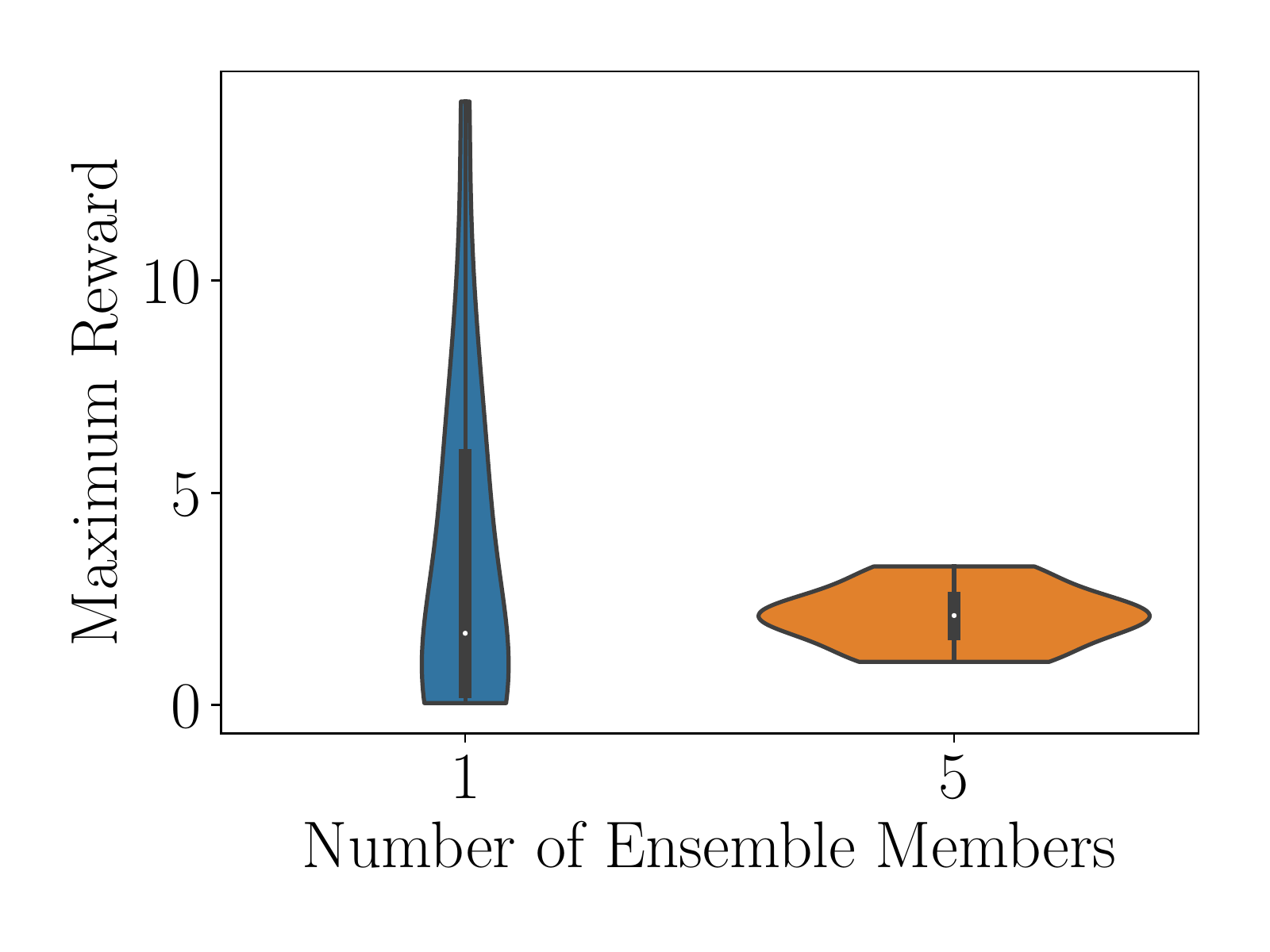}
                \caption{Max reward} 
                \label{fig:max-reward}
            \end{subfigure}
        \end{minipage}
    \end{minipage}
    \par\end{centering}
    \caption{Ensembles eliminate relearning failures in the stay inside environment.
 (b) depicts the ground truth reward in the stay inside environment. (c) shows an example individual learned reward and (d) with a five member ensemble. Finally, (e) shows the distribution of max learned-reward across all states. All sub-figures come from the same run which included 20 seeds.}
    
    \label{fig:reward_structure}
    \vspace{-0.5cm}
\end{figure}

} \fi

Our tabular experiments provide a concrete, interpretable example of how relearning failures can be effected by reward model implementation details. In particular, we focus on reward ensembles and observe that they have drastic effects on relearners but leave the sampler's performance unchanged.

In the stay inside environment, when using a reward ensemble of size five, all relearners preform at least as well as their respective samplers, as can be seen in \autoref{fig:relearner-and-sampler}. However, if we use only a single reward, the relearners behaviour is inconsistent; some relearners do substantially better then their respective samplers, but almost as many do substantially worse, getting near zero return. Thus, while adding an ensemble has a minimal effect on the sampler, it changes the behaviour of the relearners.

To understand why this happens we must consider the off-distribution behaviour of our reward models. In the stay inside environment, the samplers typically stay in the inside half of the environment. Thus there is often insufficient coverage of the outside half of the environment in our trajectory dataset. Thus, the reward off-distribution is largely unconstrained by the data. This means that small changes in the off-distribution behavior of our reward network can become critically important. Reward models based on neural networks  produce spurious high rewards off distribution, see \autoref{fig:reward_structure}. When these \textit{reward delusions} are more rewarding than any of the in-distribution transitions, reward hacking can occur and cause relearning failures. Reward ensembles tends to have lower variance off distribution than an individual reward network. Thus, any reward delusions tend to have a lower reward (according to the reward model). This can be directly seen in \autoref{fig:reward_structure} (c-e). This effect reduces the chance that the optimal policy will be attracted to one of these spurious rewards during relearning, which is what we see in \autoref{fig:relearner-and-sampler}.

\section{Limitations and Discussion}

Our experiments have a few important limitations. First, they are limited to simple ground truth reward functions and environments. For example, in \texttt{Half-Cheetah-v3} \citep{openaiGym, seals2020}, the reward function is essentially a linear in the observation, action and next observation. While these relearning failures also appear in more complex tasks \cite[Section~3.2]{ibarzRewardLearningHuman2018}, it is unclear if it is precisely the same phenomenon that causes them. The design decisions that seem to improve retraining performance in small-scale experiments, in our case, reward ensembles and less sampler training, may not be the same as those that address the problem at a larger scale. We leave such explorations to future work.

Overall, we have demonstrated that evaluations of relearning performance can differ substantially from the results of simply evaluating the sampler agent trained alongside the reward model. We hope to see future works include relearning evaluation as they appear to hold fruitful insights into the quality of the learned reward functions.
\newpage

\section*{Author Contributions}

Lev McKinney designed and implemented the tabular experiments and wrote the relevant parts of the method and experiments sections. In addition, he wrote the introduction, discussion and related works sections of the paper/appendix. Yawen Duan designed and ran the initial experiments that displayed reward model relearning failure on continuous control environments, and wrote relevant sections of the paper. David Krueger provided ideas, guidance and general feedback on experiment design and analysis. Adam Gleave provided initial ideas of the project, provided high-level and detailed feedback on experiments and analysis.

\begin{ack}
This paper was completed as part of an internship at the Center for Human-Compatible Artificial Intelligence. Funding for this internships was provided by the Berkeley Existential Risk Initiative.
\end{ack}

\bibliography{references}
\begin{appendices}
\section{Training Details and Hyperparameters}
\label{appendix:hyperparameters}

\subsection{Reinforcement learning algorithms}
In the tabular setting we train the sampler policy using soft Q-Learning \cite{haarnojaReinforcementLearningDeep2017}. We use soft actor-critic (SAC) \citep{haarnoja2018sacapps} implementations of Stable-Baselines3 ~\citep{stable-baselines3} in the locomotion control tasks.

Both algorithms are off-policy and use a replay buffer, which ensures their high sample efficiency compared to on-policy RL algorithms. Note that the learned reward function $\hat r_\phi$ changes during training, so we relabel the transitions in the replay buffer after each iteration, similar to PEBBLE \cite{leePEBBLEFeedbackEfficientInteractive2021}. The main difference between our algorithm and PEBBLE is that we omit the unsupervised pre-training stage used in PEBBLE. We used the implementations from Imitation Learning Baseline Implementations ~\citep{wang2020imitation} to perform the experiments.

\subsection{Continuous Control Experiments}
In the tabular setting, all reward networks only take the current state as a one-hot vector. They consist of a multi-layer perceptron with two hidden layers of size 256 and ReLU activations, similar to those used in PEBBLE \citep{leePEBBLEFeedbackEfficientInteractive2021}.

\paragraph{Training details} For reward learning experiments, we used the implementations of Preference Comparisons Algorithm from Imitation Learning Baseline Implementations ~\citep{wang2020imitation} with a full list of hyperparameters in Table \ref{tab:pc}. For the RL component, we used soft actor-critic (SAC) \citep{haarnoja2018sacapps} implementations from Stable-Baselines3 ~\citep{stable-baselines3} in the locomotion control tasks with a list of hyperparameters in Table \ref{tab:sac}. For retraining evaluations, we use the same hyperparameters for SAC to train new agents against the frozen learned reward models.

\paragraph{Reward model} The reward model consists of a single multi-layer perceptrons with two hidden layers of size 256 and LeakyReLU activations with slope 0.01. The input of the model consists of the state, action and next state vectors, and the input vector is normalized by running normalization. The output the the reward model is normalized by by exponential moving average. During relearning experiments, we directly use the raw reward output from the reward network while being normalzed by a VecNormalize layer in Stable-Baselines3 (\texttt{https://stable-baselines3.readthedocs.io/en/master/guide/vec\_envs.html\#vecenv}).

\paragraph{Reward normalization} We compute a normalized version of the learned reward using an Exponential Moving Average to normalize the reward to mean zero and unit standard deviation. This normalized reward was then used for policy optimization. Note that normalizing the reward does not change the optimal policy, which is invariant to positive affine transformations. However, it does simplify the optimization problem. In particular, a normalized reward is a more stable objective for the critic to learn over time. Additionally, RL hyperparameters can depend on the reward scale (for example, learning rate should be set inversely proportional to reward scale) -- normalizing the learned reward therefore allows us to use a consistent set of hyperparameters.

\begin{table}[!ht]
    \centering
    \begin{tabular}{|c|c|}
        \hline
        Hyperparameter  & Value \\ \hline \hline  %
        Segment Length  & 50    \\ \hline
        Total Comparisons & 2000 \\ \hline
        Number of Iteration & 50 \\ \hline
        Reward Training Epochs & 5 \\ \hline
        Query Schedule & constant \\ \hline
    \end{tabular}
    \caption{Reward learning hyperparameters for continuous control experiments}
    \label{tab:pc}
\end{table}

\begin{table}[h!]
    \centering
    \begin{tabular}{|c|c|}
        \hline
        Hyperparameter  & Value \\ \hline \hline  %
        Learning Rate  & 0.0003    \\ \hline
        Batch Size  & 256    \\ \hline
        Discount & 0.99 \\ \hline
        Learning Starts from & 10000 \\ \hline
    \end{tabular}
    \caption{SAC hyperparameters for continuous control experiments}
    \label{tab:sac}
\end{table}

\subsection{Tabular Experiments}
Similarly to the continuous control experiments we use Imitation's implementation of preference comparison \citep{wang2020imitation}. However, we use a tabular soft-q learning algorithm with a replay buffer \cite{haarnojaReinforcementLearningDeep2017} with reward relabling \citep{leePEBBLEFeedbackEfficientInteractive2021} to solve the environments. The reward network again uses a similar MLP architecture to the continuous control setting with a sightly smaller hidden size of 32. Finally, we normalize the reward functions before ensembling them using a simple running norm over sampled transitions which is frozen during retraining. Hyperparamaters can be found in \autoref{tab:tabular-hyperparameters}.

\begin{table}[!ht]
    \centering
    \begin{tabular}{|c|c|}
         \hline
         Hyper Parameter & Value \\ \hline \hline
         \multicolumn{2}{|c|}{Sampler Soft-Q Learning} \\ \hline
         discount & 0.99 \\ \hline
         learning rate & 5e-2 \\ \hline
         replay buffer capacity & $\infty$ \\ \hline
         temperature & 0.1 \\ \hline
         samples from buffer per env sample & 10 \\ \hline
         initial soft-q value & 200 \\ \hline
         \multicolumn{2}{|c|}{Reward Learning} \\ \hline
         trajectory fragment length & 30 \\ \hline
         total comparison budget & 2,500 \\ \hline
         RL budget & 500,000 \\ \hline
         frac. of comparisons from inital random trajs & 0.1 \\ \hline
         select fragments for comparison & randomly \\ \hline
         epochs of training per iteration & 1 \\ \hline
         number of iterations & 100 \\ \hline
         query schedule & constant \\ \hline
         reward learning rate & 1e-3\\ \hline
         \multicolumn{2}{|c|}{Reward Network} \\ \hline
         reward network hidden layers & [32, 32] \\ \hline
         activation function & ReLu \\ \hline
         output normilization & Running Norm \\ \hline
\end{tabular}
    \caption{Tabular Experiment Hyperparamerers}
    \label{tab:tabular-hyperparameters}
\end{table}

\paragraph{Tabular Relearning} When relearning we solve for the soft-optimal policy under the learned reward function with temperature 0.1 and discount factor 0.99.

\iflongversion\else{
\section{Environments}
\paragraph{Locomotion Control Task}

We ran reward learning and relearning on a MuJoCo locomotion task~\citep{todorovMujoco2012} -- \texttt{HalfCheetah} environment from the \emph{seals} benchmark suite~\citep{seals2020}, a modification of \texttt{HalfCheetah-v3} in the \textit{gym} environment suite which adds the x-coordinate of the robot’s center of mass (COM) to the first dimension of the observation space. The ground-truth reward function of the \texttt{HalfCheetah} environment is a linear combination of the x-velocity of the robot’s COM and a control cost dependent on the $L_2$ norm of the action vector. Consequently, the reward function in \emph{seals} \texttt{HalfCheetah} is a function of the observations, which is not strictly true in the original \textit{gym} \citep{openaiGym} environment, avoiding a potential confounder.

\paragraph{Tabular Environment}

We constructed the \textbf{stay inside environment}, which consists of a 20x20 closed grid of cells. The top "outside'' and bottom "inside'' halves of the environment are separated by a wall with a narrow two cell gap in the middle. The reward for each state is shown in \autoref{fig:reward_structure} (a), with reward values ranging from +10 to -1.
}\fi

\section{Epic Distance as an Evaluation Metric}
As an additional evaluation criterion, we consider using EPIC distance \cite{gleaveQuantifyingDifferencesReward2021} to measure the distance between learned reward functions and the ground truth reward. EPIC works by canonicalizing the rewards to be invariant to potential shaping, normalizing them to be invariant to scale, and then computing the $L^2$ norm of the difference of those functions over a \emph{coverage} distribution of transitions. Here we consider two coverage distributions: uniform and expert distribution. The uniform distribution is uniform over feasible transitions. The expert distribution is the distribution of a soft-optimal policy with a temperature of 10 to give slightly more coverage.

\section{Additional Tabular Experiments}
\begin{figure}
    \centering
    \begin{tabular}{cc}
         \includegraphics[trim=0 0 0 0, clip, width=0.47\textwidth]{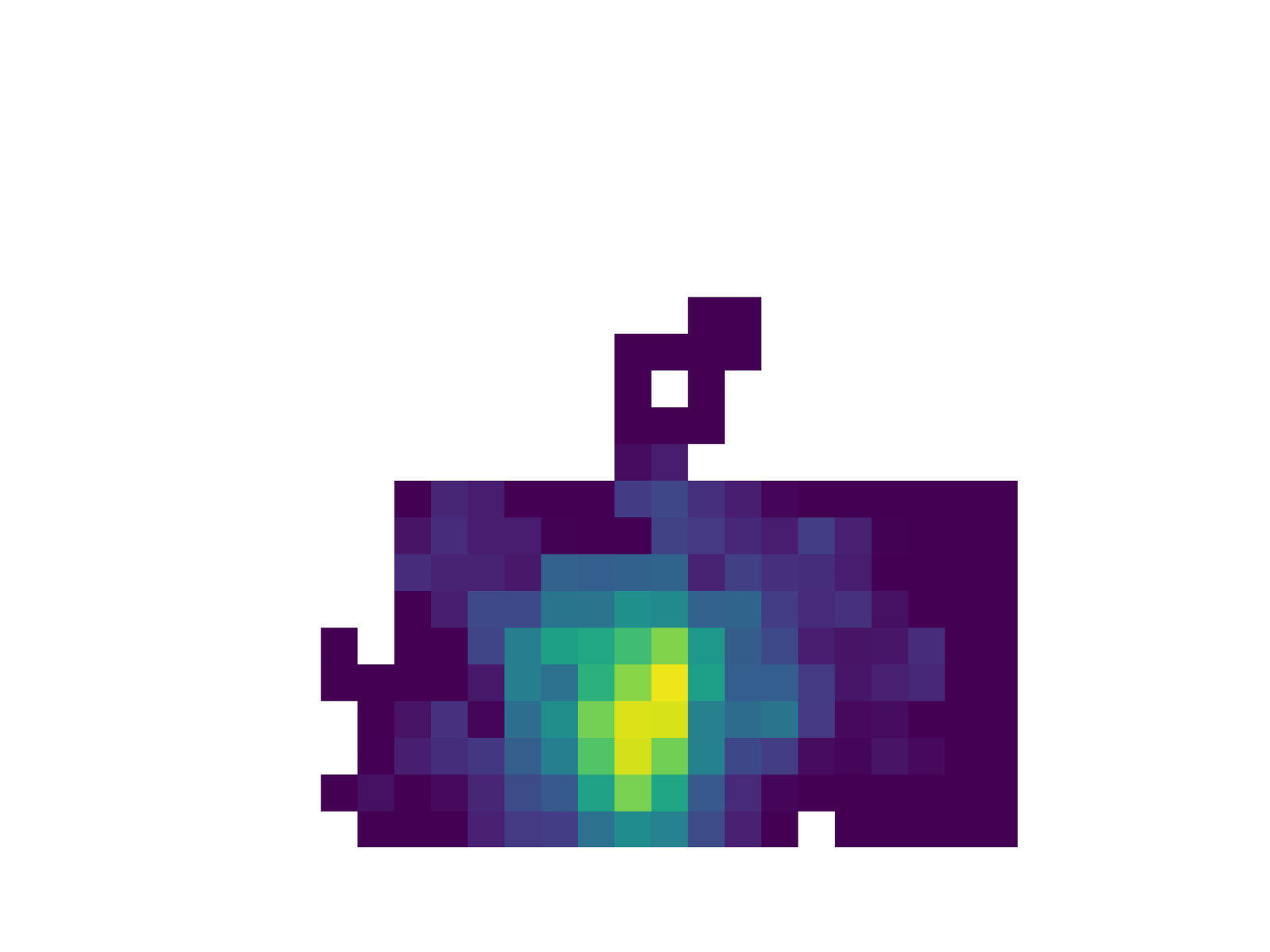}&
         \includegraphics[trim=0 0 0 0, clip, width=0.47\textwidth]{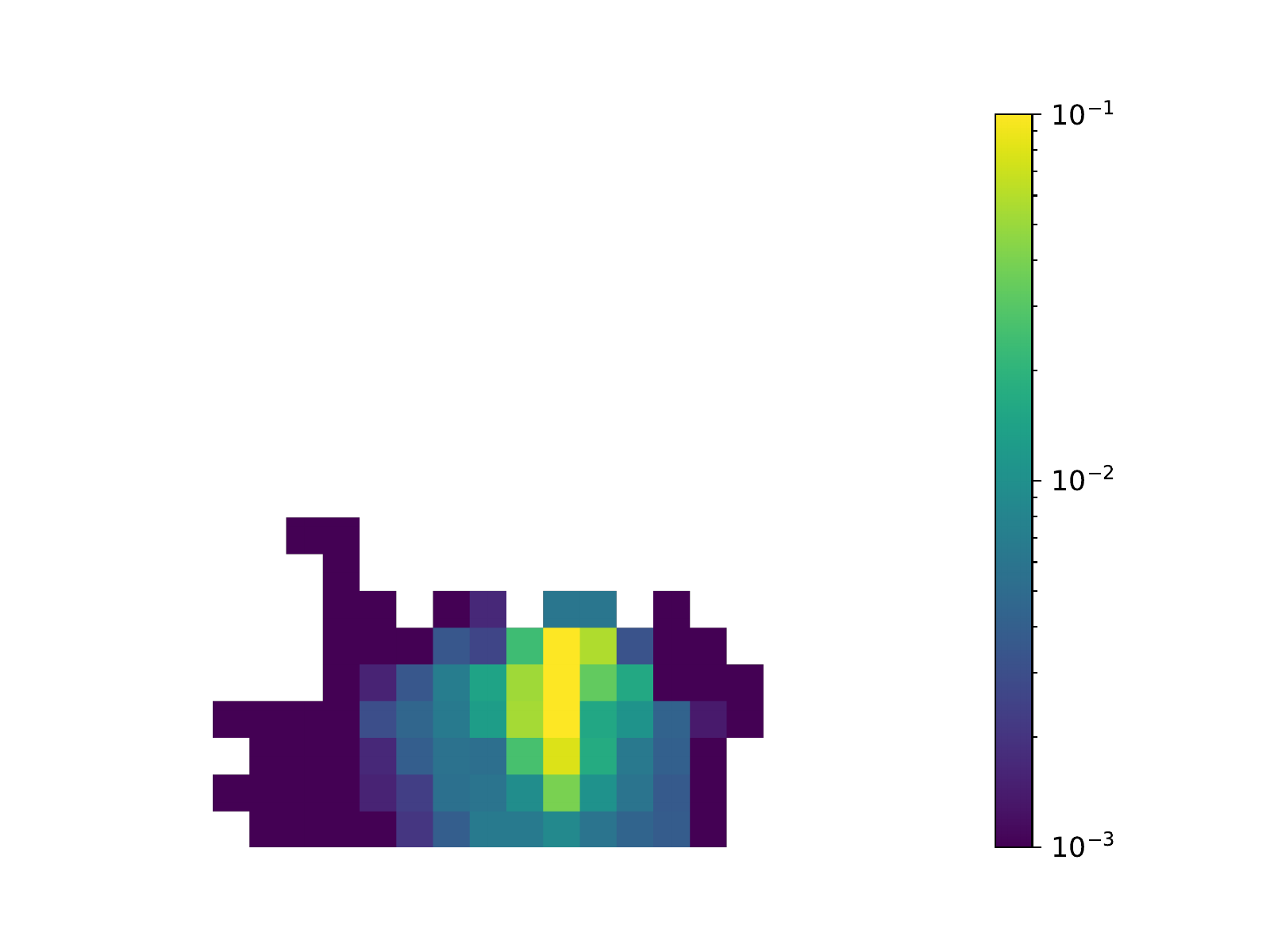}\\
         (a) Example without ensemble & (b) Example with ensemble
    \end{tabular}
    \caption{Example on policy distribution}
    Examples of the on policy distributions of the samplers in the stay inside environment, marginalized over the entire training run.
    \label{fig:on-policy}
\end{figure}

\label{appendix:addition_tabular}
\begin{figure}[!ht]
    \begin{center}
    \begin{tabular}{cc}
         \includegraphics[width=0.5\textwidth]{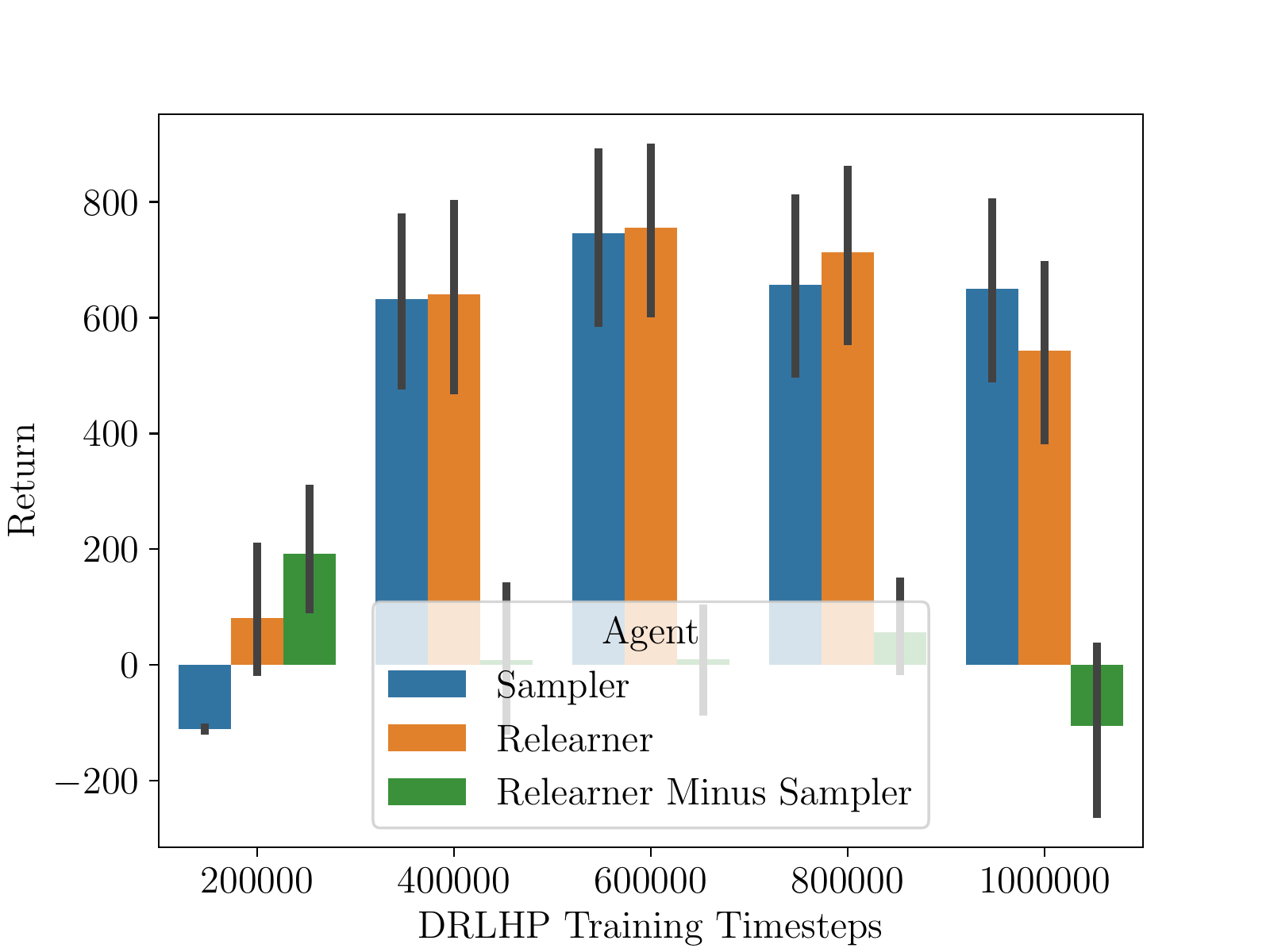} & \includegraphics[width=0.5\textwidth]{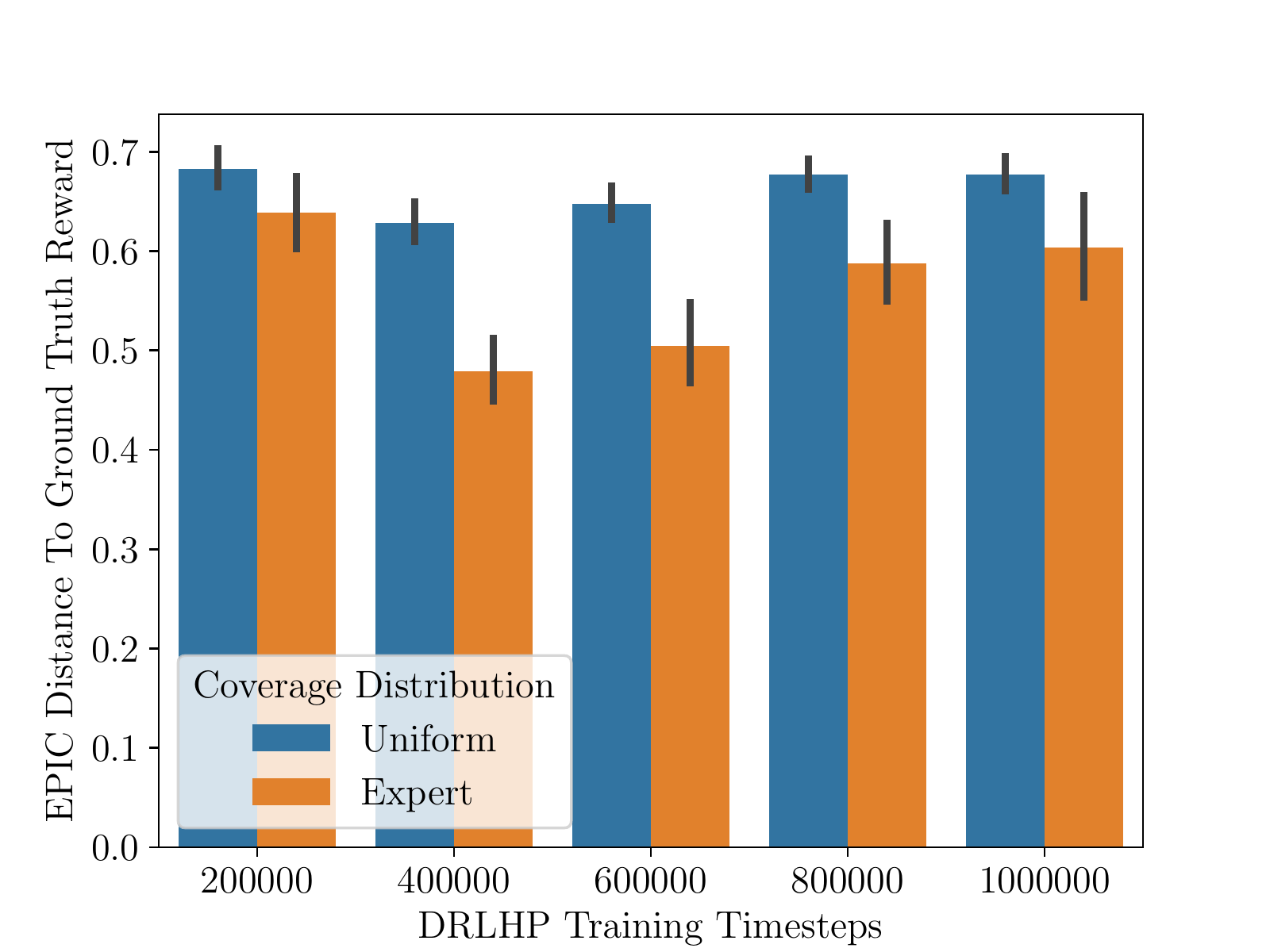} \\
         (a) & (b)
    \end{tabular}
    \par\end{center}
    \caption{Increasing the number of time steps of R.L. training does not seem to significantly effect relearning failures.}
    \label{fig:tabular-timesteps}
\end{figure}

\begin{figure}
    \begin{center}
    \includegraphics[trim=10 0 10 0, clip, width=0.5\textwidth]{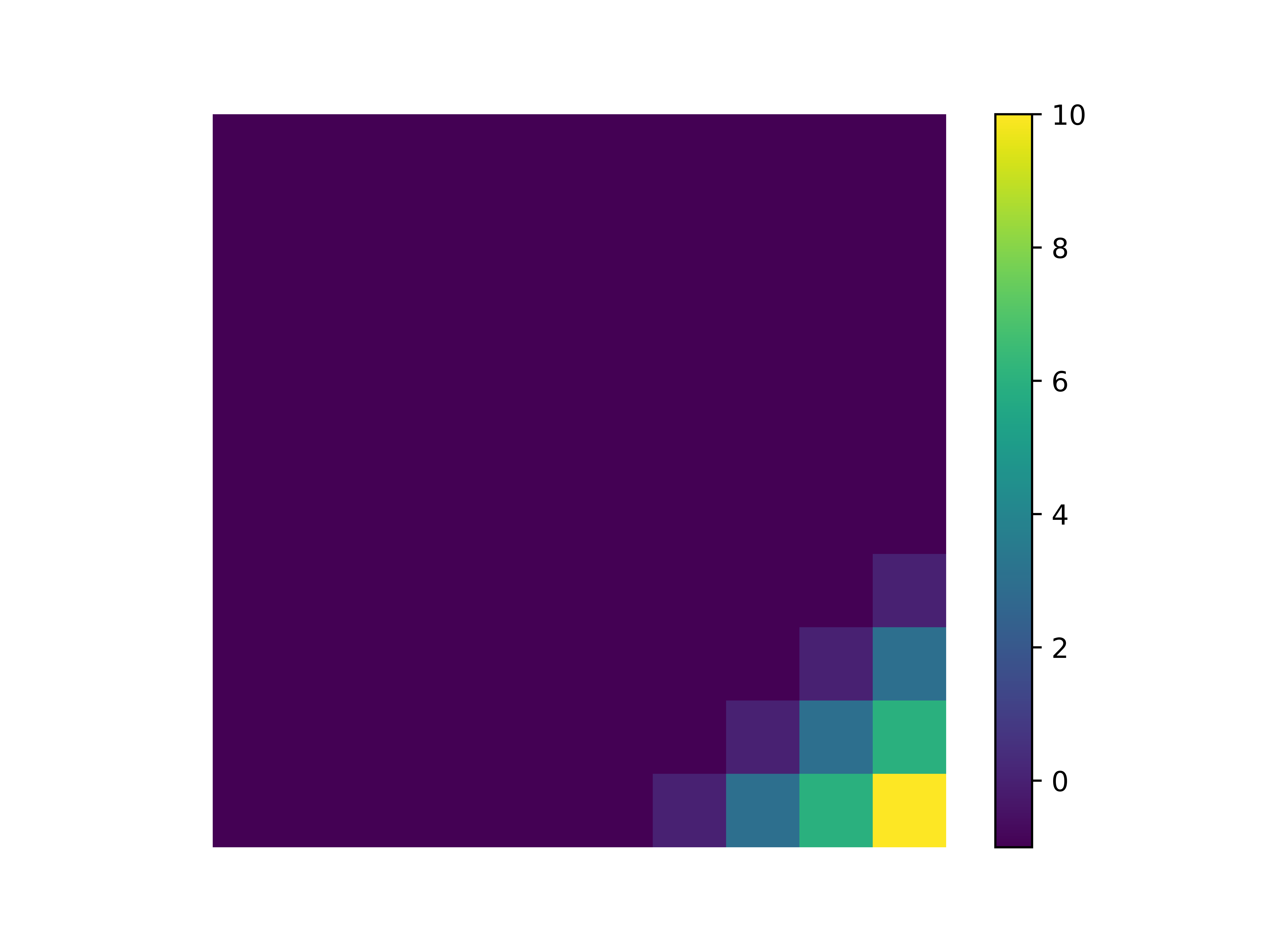}
    \par\end{center}
    \caption{Tiny room environment. The ground-truth reward in the tiny room environment. Note that the reward only depends on the current state.}
    \label{fig:reward-tiny}
\end{figure}

To study the effects of training the sampler for a more time steps, we first consider a simple environment consisting of a 10x10 grid world. The agent begins in the lower left-hand corner of the environment and gains a ground-truth reward of 10 for reaching the lower right-hand cell, as seen in \autoref{fig:reward-tiny}. 

The performance of the sampler and relearner initially increases with more training timesteps, with our relearners generalizing well and achieving slightly higher performance than their respective samplers. However, it quickly plateaus even though we \textit{do not} see significant reductions in relearner performance with an increased number of time steps. The EPIC distances of our learned reward functions from the ground truth reward begin to increase after 400,000 timesteps \autoref{fig:tabular-timesteps} (b).

Increasing the number of total training timesteps used for DRLHP does seem to degrade the quality of the reward function according to EPIC distance. However, it does not appear to hurt relearning performance in the same way in this simple tabular environment.

This is a strikingly different effect than we see in \texttt{HalfCheetah}. This may be because in a tabular setting the sampler either finds the optimal policy induced by the learned reward function every iteration, so the sampler and relearner have equal performance, or it insufficiently explores the environment, and reward learning completely fails. This dichotomy leaves little room for the subtle degradation in relearner performance we see in \autoref{fig:longer}.
\end{appendices}
\end{document}